# On Fixing the Right Problems in Predictive Analytics: AUC Is Not the Problem

Ryan S. Baker[1], Nigel Bosch[2], Stephen Hutt[3], Andres F. Zambrano[1], Alex J. Bowers[4]

1. University of Pennsylvania

2. University of Illinois Urbana-Champaign

3. University of Denver

4. Teachers College Columbia University

Recently, ACM FAccT published an article by Kwegyir-Aggrey and colleagues (2023), critiquing the use of AUC ROC in predictive analytics in several domains. In this article, we offer a critique of that article. Specifically, we highlight technical inaccuracies in that paper's comparison of metrics, mis-specification of the interpretation and goals of AUC ROC, the article's use of the accuracy metric as a gold standard for comparison to AUC ROC, and the article's application of critiques solely to AUC ROC for concerns that would apply to the use of any metric. We conclude with a re-framing of the very valid concerns raised in that article, and discuss how the use of AUC ROC can remain a valid and appropriate practice in a well-informed predictive analytics approach taking those concerns into account. We conclude by discussing the combined use of multiple metrics, including machine learning bias metrics, and AUC ROC's place in such an approach -- like broccoli, AUC ROC is healthy, but also like broccoli, researchers and practitioners in our field shouldn't eat a diet of only AUC ROC.

CCS CONCEPTS • Computing Methodologies → Machine Learning • Social and Professional Topics → Computing and Society → Ethics• General and Reference → Cross-computing Tools and Techniques → Evaluation

**Additional Keywords and Phrases:** AUC ROC, predictive analytics, model validity

## 1 INTRODUCTION

Predictive analytics have become a large part of our contemporary world. Machine-learned models are now used at scale for a large range of applications, from predicting high school student dropout [36] to credit default [42] to prison recidivism [59]. A wide range of both theoretical and applied research has emerged to study the properties of different algorithms and metrics [4, 26] and key issues in their application in a range of fields [57, 25, 22]. One concern that has repeatedly (and appropriately) emerged is the fairness of these algorithms and their application [39, 40, 29].

In our paper, we consider a recent article about inappropriate and appropriate practice in predictive analytics, "The Misuse of AUC: What High Impact Risk Assessment Gets Wrong", published by Kwegyir-Aggrey and colleagues at the most recent iteration of ACM FAccT [33]. This paper reviewed and discussed a number of important challenges in the use of predictive analytics, offering a central framing that the use of the AUC ROC metric [41, 23] is a core problem in this area. If correct in its central claims, this paper would offer a major challenge to standard practice in the field, as AUC ROC is one of the most -- perhaps the most -- widely used classification metric in many application areas of predictive analytics.

In our paper, we offer a critique to several of Kwegyir-Aggrey and colleagues' claims. Many of the problems the authors bring up are substantive, significant, and need to be addressed. However, we dispute one of the central claims of their article: that the source of the problems the authors raise is the use of AUC ROC. Many of the problems that the authors identify are not problems inherent to the use of AUC ROC, but could pertain to the use of any metric in a singular fashion.

In this current paper, we will also discuss technical areas where the authors' statements and interpretations about AUC ROC do not match commonly held and well-documented findings about this metric, such as class imbalance, and where easy-to-apply solutions already exist for the problems the authors identify, such as some parts of the model threshold space being of more importance than others. These issues are important to explicate and debate, to drive fuller community understanding of how AUC ROC functions, both effectively and ineffectively.

The core of our concern is that AUC ROC, for its flaws, is a metric with many virtues and uses. Kwegyir-Aggrey and colleagues title their paper as being about the "misuse" of AUC ROC, but in the article itself repeatedly appear to call for the elimination of use of this metric. Eliminating the use of AUC ROC without a better solution ready-at-hand will not fix the problems in the field. In places, the authors seem to argue for replacing AUC ROC with the metric typically called "accuracy" (i.e., proportion correct), comparing accuracy's conclusions with AUC ROC's, and using the difference between these metrics as evidence that AUC ROC is flawed. As we will discuss below, there is a long history of critique of accuracy, and replacing AUC ROC with accuracy has considerable risk for worsening outcomes for the individuals impacted by predictive analytics.

Overall, we concur that there are significant flaws in the current use of predictive analytics in practice, in terms of fairness, accountability, and transparency, that need to be addressed by the field. However, AUC ROC is not itself the problem. In the remainder of this document we will respond to this paper's critiques of AUC ROC, in order to highlight our views on where the problems and challenges lie for the field, and caution the field away from the intemperate abandonment of AUC ROC, a step that will exacerbate problems rather than addressing them.

## 2 CRITIQUES THAT APPLY TO ANY METRIC

One of the limitations in the Kwegyir-Aggrey et al. paper is that it applies several critiques to the use of AUC ROC that could be applied to any metric. For example, the authors state that "practices involving AUC are not robust, often invalid, and may hide policy choices under the guise of technical development" and cite Singh et al. [51] as identifying that AUC ROC is misinterpreted in many papers published on recidivism prediction.

We would concur that many papers do not use AUC ROC in robust or valid ways -- but this is not so much a flaw with AUC ROC as with the often-careless way that practitioners use many metrics, not just AUC ROC. For example, many papers could be found in the literature which ignore the serious base rate concerns with the metric accuracy, as Kwegyir-Aggrey et al. themselves do (see [53, 54] for a discussion of this issue). Concerns have also been raised with the common uses of kappa and the F1 statistics (see [13, 10] for a discussion of this issue) in predictive analytics, and -- for that matter -- with the uses and interpretations of statistical significance testing in the broader literature [38]. We also concur that a single-minded focus on any metric, by itself, and selection of models based on performance on a single metric is and has long been seen as problematic [46]. However, as the authors themselves cite in referring to Singh et al. [51], many of the papers that use AUC ROC to predict recidivism do present other metrics that present a more nuanced story than AUC ROC alone. This use of multiple metrics is common in other areas as well, such as educational prediction [36, 34].

Kwegyir-Aggrey et al. also mention, in discussing Perdomo et al. [44], that models may be able to predict future outcomes successfully in general, but may be biased. In that example, they also note that despite effectively predicting future outcomes, the interventions based on those models were ineffective, and that educators were not trained in how to use the dashboard. These are unfortunate outcomes, and all represent serious problems. There has fortunately been considerable work on studying and debiasing algorithms used in risk prediction (see reviews in [3, 31]) and in trying to design better interventions for high school dropout [50, 55] and better approaches to teacher training [30]. However, it is unclear how these problems are due to the use of AUC ROC (or any model goodness metric). Each of these issues is important to address, and is separate from whether AUC ROC is used. Algorithmic bias should be considered as well as overall algorithm predictive performance. No algorithm is much use if it is embedded in ineffective or inappropriate interventions. Finally, few interventions function well if the practitioners who will use them are not offered professional development and/or training. These challenges can be addressed (or not addressed), regardless of which metric is selected to evaluate overall model performance.

## 3 DEFINITIONAL AND TECHNICAL ISSUES

### 3.1 What do we mean when we talk about model "accuracy"?

The term "accuracy" comes up 50 times in [33], and is key to several of the critiques of AUC ROC seen in the paper. For example, on page 1570, the authors refer to AUC ROC "as a replacement for model accuracy evaluation"; on page 1572, they state that their goal is to "dispute the common assumption that AUC can be used to measure predictive accuracy" and explicitly state that one of their goals is to "explain how the two metrics are correlated" and that "AUC based measurements of predictive accuracy can be noisy and inaccurate." Indeed, an entire subsection (4.1) is devoted to showing how AUC ROC and "accuracy" disagree, treating disagreements as evidence against the validity of AUC ROC. Throughout section 5, the authors also critique practitioner and governmental statements that refer to high AUC ROC as demonstrating accuracy.

Part of the challenge to thinking about AUC ROC and accuracy is that the term "accuracy" is used in two distinct ways by researchers and practitioners in this field. First, there is a technical definition used within the field of machine learning and subfield of predictive modeling. In that community, "accuracy" is a specific metric used to evaluate the performance of a classification model. It is defined as the percentage of correctly classified instances out of the total instances. In mathematical terms, it is the ratio of the sum of true positive and true negative predictions to the total number of predictions. This measure provides a quick and straightforward assessment of a model's performance, perhaps most interpretable in scenarios where the classes are balanced, and least interpretable in cases of substantial data skew [28].

However, outside the specialized domain of machine learning, the term "accuracy" often carries a broader and more generalized meaning. Stakeholders not deeply versed in machine learning algorithms (such as government officials, funders etc.) might use "accuracy" to describe the overall effectiveness or success of a model at predicting what it is intended to predict. This broader usage of this term encompasses various other performance metrics, such as precision, recall, AUC ROC, F1, Kappa, and others. Success is typically highly context-dependent, and so the exact metric being used to define success may vary. However, it is not uncommon for stakeholders to use the term "accuracy" in referring to this more general notion of model success. In documents written for non-experts, or for a general popular audience, it is therefore typical to see this more broad definition of accuracy used, as a general audience may not know the technical terms AUC ROC, F1, etc.

For example, on page 1578 of the Kwegyir-Aggrey et al. article [33], they critique an IES technical report [6] for conflating AUC ROC with accuracy. However, in the footnote to the very quote that Kwegyir-Aggrey et al. [33] critique, Bruch and colleagues state "The strength of predictions, or accuracy, is measured using a metric called the area under the curve (AUC)" -- indicating that they are using "accuracy" in a broader sense. In general, Bruch et al. [6] are writing for a general audience and increasing the accessibility of their article by using the broader definition of accuracy, rather than the specialized definition researchers are used to seeing in machine learning.

## 3.2 Should the metric "accuracy" be used as a gold standard?

In section 4.1 of their article, Kwegyir-Aggrey et al. [33] compare AUC ROC and accuracy, repeatedly noting where the two metrics diverge and using these divergences as evidence of AUC ROC's invalidity. This critique would be essential if many in the field thought that AUC ROC and accuracy communicated the exact same information about a model. However, given that the term "accuracy" is used to mean two very different things, this critique becomes more problematic. As Kwegyir-Aggrey et al. [33] note in their article, but then ignore henceforth, there are many serious problems with the metric named accuracy, well-known to the field. Treating the metric named accuracy as a gold standard to evaluate other metrics by (as in section 4.1 of [33]) is a problematic choice, given accuracy's susceptibility to class imbalance, equal weighting of false positives and false negatives, and lack of consideration of model confidence. There are a large number of papers and textbooks criticizing accuracy -- see, for instance [27, 28, 56, 17, 2, 54].

## 3.3 Further notes on class imbalance

Another of Kwegyir-Aggrey et al.'s critiques is that AUC ROC handles class imbalance poorly. Kwegyir-Aggrey et al.'s justification for this claim, found in section 4.1.2 of their paper, appears to be that AUC ROC has higher confidence intervals when classes are imbalanced than when they are balanced. We concur that this is true, but question this as a critique of AUC ROC, for two reasons.

First of all, with AUC ROC it is actually possible to calculate these confidence intervals in a straightforward, mathematical fashion, and to assess the degree to which class imbalance impacts confidence intervals. This is much more difficult for other metrics; for example, Kwegyir-Aggrey et al.'s preferred accuracy does not have a closed-form mathematical way to calculate confidence intervals. Instead, the confidence intervals of accuracy must be calculated on a case-by-case basis by using bootstrapping or other computationally intensive procedures, and these estimates are divergent and often unreliable [60].

Furthermore, past comparisons of reliability of estimation between metrics have suggested that AUC ROC is more robust to class imbalance than accuracy and several other popular metrics [28, 54, 35]. In other words, other metrics are more impacted by class imbalance than AUC ROC. Kwegyir-Aggrey and colleagues cite Lee & Chung [34] as evidence for their claim that class imbalance is a major problem for AUC ROC, but that appears to be a mis-citation; that paper is on a different topic, how to improve model performance by addressing class imbalance. In general, AUC ROC's handling of class imbalance is typically considered by the field to be a strength rather than a weakness.

### 3.4 Cut-offs for AUC ROC values

Kwegyir-Aggrey et al. [33, p. 1578] also assert that one of the core goals of using AUC ROC is to establish a single performance cutoff point above which a model is deemed to have "strong model fit," which warrants careful scrutiny. We agree that the selection of a default threshold as "good enough" is a problematic practice. We concur that relying on a universal "good enough" threshold, such as an AUC ROC value above 0.7, is rarely advisable. Machine learning models are typically deployed in diverse and complex real-world scenarios where the stakes and consequences of predictions can vary significantly. Different applications demand different trade-offs between precision (the proportion of true positives among all positive predictions) and recall (the proportion of true positives among all actual positives). For instance, in medical diagnostics, a high recall might be prioritized to ensure all potential cases are identified, even at the cost of precision. A universal cut off would not support this, and is thus not considered best practice.

While the use of AUC ROC often facilitates comparison between models, due to the universality of meaning of AUC ROC, even across different data sets [18], and chance level of 0.5 for all data sets (properties not present for accuracy), there is no good justification for selecting a value a priori as "good enough" across all data sets. Instead, best practices in machine learning emphasize the need for a contextual, nuanced approach to model evaluation, where the characteristics and requirements of each application guide the interpretation of performance metrics.

However, it is worth asking how common this practice of asserting an a priori cut-off for AUC ROC is. Kwegyir-Aggrey et al. [33] find one paper by a practitioner that does so, but also claim that Rice and Harris [48] recommend an a priori AUC ROC cutoff. A thorough review of that paper yields no such recommendation. One technical report using AUC ROC in this flawed fashion -- or even several -- is not justification for discarding AUC ROC as a metric.

### 3.5 What is AUC ROC trying to do?

On page 1571, Kwegyir-Aggrey et al. [33] claim that the goal of AUC ROC is that "the AUC attempts to measure the degree to which the classifier can produce a distinct decision point (represented by a single $\lambda$) that achieves a good trade-off between true and false positives. The larger the AUC, the more well defined this decision point will be."

We are not familiar with this definition of AUC ROC's goals, which the authors offer no citation for. Other works discussing AUC ROC [41, 23] typically instead view AUC ROC as evaluating model performance at every possible threshold and considering different thresholds equally. There are clear limitations in that approach; as the authors point out elsewhere, this approach does not consider the very real possibility that some parts of the decision space matter more than others. But the goal of measuring performance across thresholds is very different than finding the optimal decision point; a practitioner or researcher with the goal of finding and evaluating the optimal decision point would be better served by -- for instance -- searching the space of thresholds to find the threshold with the highest F1 or Kappa.

Kwegyir-Aggrey et al. [33] also discuss model calibration (referred to as "fit" in their paper), or the degree to which a model's probability predictions reflect the actual probability of an outcome. For example, out of all predictions P(Y1) ≈ .75 for outcome Y1, about 75% of those predictions should be correct (i.e., Y1); if more are correct, the model is under-confident at the .75 threshold, while if fewer are correct, the model is over-confident. Good calibration can be a useful property for models, since it allows for straightforward selection of decision thresholds that have a particular precision, though it is also possible to do so with an imperfectly calibrated model, e.g., by examining precision–recall curves. AUC ROC does not measure or favor models with good calibration, which Kwegyir-Aggrey et al. [33] refer to as "crucial issues". However, this same limitation is true of many metrics, including all threshold-based metrics such as accuracy. Caruana & Niculescu-Mizil [7] show that only squared error, cross-entropy, and CAL (a metric that exclusively measures calibration) appropriately capture calibration information. If such information is critical, it is highly appropriate to use a metric like CAL to supplement interpretation about a model, regardless of whether performance is primarily measured via AUC ROC, accuracy, or another metric

## 4 MODEL CHOICE

A common step in machine learning development processes is selecting a particular model from among many choices [1, 49, 47]. For example, there may be choices between learning algorithms (e.g., random forest versus support vector machines), hyperparameters to tune (e.g., number of layers in a neural network, depth of decision trees), or features to select based on which yield the most performant model. Each of these model selection tasks requires a metric that defines how good a model is, and thus—via sorting—which is best. AUC ROC is one possible metric for this step, which, like all metrics, is suitable only in certain cases. As Kwegyir-Aggrey et al. [33] note -- and we agree -- selecting a model according to AUC ROC must be done with care.

If the goal is to select a model that works well for one specific decision threshold (i.e. cut-off, where students on only one side of the cut-off receive the intervention), it is perhaps more natural to choose a metric that measures performance at that threshold. For example, if predicting whether or not a student will pass a particular course [11, 52, 45], the performance of a model at a threshold where it predicts that 90% of students fail the course is likely not relevant to the utility of that model, whereas its performance at a threshold that results in 5% failure predictions might be far more relevant. AUC ROC may seem inappropriate for such cases. However, even in such cases the *partial* AUC ROC [15, 37] could be of value for two reasons. First, many machine learning decision systems serve as probabilistic advice to human decision makers rather than as categorical decision-making systems [12, 21]. In these cases, while not all possible thresholds are often of interest, the performance across a more limited range of thresholds often is. Second, even when only one specific threshold is of interest, the performance in the neighborhood of that threshold may be worth considering, given that real-world data variation and distribution shift imply that a model that works well across a range of thresholds is likely preferable to one that is only known to work well at one threshold in the dataset on which it was trained/tested. Additionally, in many common cases, two or more interventions are feasible and have different cost-benefit trade-offs. In this situation, a metric tailored to a single threshold may be non-optimal.

Moreover, while it might be conceptually compelling to visually compare models across all values of the ROC curves rather than "only comparing AUCs without comparing the ROC curve from which they came" [33], it is important to note that selecting the best model requires a metric that can sort (or rank) models such that there is a best model. Selecting a single model based on multiple values—whether they are the values of points on the ROC curve or the values of multiple metrics such as F1, RMSE, and others—is not possible except in the trivial case where all metrics agree which model is the best, or through averaging together several metrics, which itself becomes a single hard-to-interpret metric. In general, comparing and sorting based on multiple values produces a Pareto front [16]: in this case, a set of models that maximize one particular metric perhaps at the expense of others. That said, it is common practice to use multiple metrics to understand one or more models (e.g., [19, 58, 9], even if not necessarily to select one. For example, Kwegyir-Aggrey et al. [33] critique the practitioner-targeted article [6] for selecting and comparing models based solely on AUC ROC, though the same authors' more technical presentation of the models [8] reveals that they actually considered nine metrics.

Algorithmic bias concerns are also critical during model selection. As Kwegyir-Aggrey et al. [33] point out, validating models based on AUC does little to ensure a model avoids issues like high performance for one group and low performance for another. However, this is true of every commonly used performance metric that does not penalize between-group performance differences. Moreover, incorporating a term in the model selection process that minimizes between-group differences based on any metric is insufficient to avoid such criticism, given that there is an inherent tradeoff between bias definitions whenever the base rates differ between groups [5, 32].

AUC ROC may even be preferable to many other performance metrics when considering fairness, given that it is common to improve model fairness by devising group-specific thresholds after model training to minimize the biases that are identified [24]. With such "post-processing" methods for reducing bias, it is essential to select a model that performs well across a range of potential thresholds—e.g., by selecting based on AUC ROC or partial AUC ROC. For the same reason, many researchers have advocated for fair model selection using fairness metrics based on the ROC curve, such as the absolute between-ROC curve area (ABROCA) which aggregates group differences in ROC curves across thresholds [20].

## 5 WHY AUC ROC REMAINS A GOOD CHOICE

Like any metric, AUC ROC is not without its limitations, as discussed extensively above. However, it has several benefits. Within this section, we discuss the many benefits that AUC ROC offers, often unique among machine learning metrics.

AUC ROC has two interpretations, both of which are potentially useful. First of all, there is the interpretation of AUC ROC as the area under the ROC curve (literally what the metric is called). Each point on the ROC curve shows the trade-off between sensitivity and specificity of the model. When expanded to the entire curve, it shows this trade-off across all possible thresholds. By taking the area under the curve, AUC ROC becomes a single comprehensive metric that provides a summary for the model's overall functioning which no other single commonly used metric offers.

As it turns out, AUC ROC can be interpreted in a second way as well. As Hanley and McNeil [23] show, given any single positive case and any single negative case, AUC ROC is equivalent to the probability of correctly deciding which is positive and which is negative. This linkage has valuable statistical properties: two models can be statistically compared to see if one model has a better fit than the other [14], or a model's fit can be compared across two different data sets [43]. Many metrics (both accuracy and Cohen's Kappa), cannot be straightforwardly compared between different data sets, if the base rates are different [53, 54]. By contrast, AUC ROC values are universal between data sets, and can be safely compared. Furthermore, the metric has a "built in" baseline comparison (an AUC ROC of 0.5 represents chance), which further adds to the ease of interpretation of this metric. For example, consider two datasets, one with a base rate of 0.3 and the other with a base rate of 0.5. A model trained on one dataset, and then tested on both, would not be straightforwardly comparable by percentage correct, or the F1 metric, due to the differences in the datasets. These confounds are somewhat addressed by AUC ROC as the results are on the same scale, with the same anchor point of 0.5 as chance.

More broadly, AUC ROC is robust to relatively high levels of skew. For example, Jeni et al. [28] show that identically performing classifiers (in terms of predictions of specific cases) can obtain radically different accuracy, F1, Kappa, and Krippendorff's Alpha if data is more or less skewed (unequal distribution of data across different classes); AUC ROC's performance remains identical across these cases. Similarly, Thölke et al. [54] use synthetic data sampled from two overlapping distributions (representing a difficult classification task) to compare different metrics across several imbalance scenarios and Machine Learning techniques. Their results show that accuracy tends to overestimate the performance of the classifiers for high levels of imbalance across all the ML techniques. In contrast, AUC ROC remains identical across all the imbalanced scenarios for several classifiers and decreases the performance estimation in the imbalanced cases for the remaining classifier instead of overestimating it.

There are, of course, still caveats and considerations to the use of AUC ROC (as there would be in any metric). Take, for example, cases where two datasets both have a high amount of skew but in opposite directions. In these cases, similar AUC ROC values can hide opposite trade-offs between sensitivity and specificity. This can make comparison of AUC ROCs more challenging and require additional context for accurate interpretation. In such situations, AUC ROC would likely still be statistically valid, but using multiple metrics (or precision-recall curves) would provide a more complete picture. However, this limitation is true of all metrics that summarize sensitivity and specificity, also including percentage correct, F1, and Kappa.

None of these benefits should be taken as justification for using only AUC ROC and no other metrics, or for ignoring the limitations that do exist for AUC ROC. The goal of our article is to demonstrate that AUC ROC is a valuable metric when used appropriately, and that many of the critiques within the article we discuss here are overstated -- or are well stated, but applicable to all performance metrics, rather than AUC ROC in particular. AUC ROC is not perfect, and should not be the only metric used to evaluate a model. Like broccoli, AUC ROC is healthy, but like broccoli, researchers and practitioners should not eat a diet of only AUC ROC.

## 6 CONCLUSIONS

In this paper, we critically examined the assertions made by Kwegyir-Aggrey and colleagues in their recent ACM FAccT publication concerning the use of AUC ROC in predictive analytics [33]. Our principal concern with this article is that the issues identified by these authors are not intrinsic flaws of the AUC ROC method per se, but rather stem from how metrics are used, and apply to many metrics if those metrics are used in isolation.

Predictive analytics are now pervasive across various domains. With this expansion comes the challenge of effectively evaluating, understanding, and applying these tools in real-world settings for a broad audience. Our paper underscores the need for a more nuanced approach in metric selection. It is not sufficient to rely solely on one metric, given that different metrics provide complementary insights, relevant to the data at hand. Furthermore, we have emphasized the necessity of addressing the communication hurdles inherent in communicating with stakeholders with varying levels of domain knowledge. It is essential that the analytical results are interpretable and actionable for a diverse range of stakeholders, not just specialists in the field – such effort for interpretability can improve some of the misconceptions raised as concerns within the Kwegyir-Aggrey et al. paper.

We have argued for the sustained relevance of AUC ROC in predictive analytics, despite some of the issues raised in Kwegyir-Aggrey et al [33]. While acknowledging the challenges they have highlighted, we believe that stopping the use of AUC ROC entirely would exacerbate, rather than relieve, these issues. Instead, a more balanced approach, integrating AUC ROC with other metrics, including fairness metrics, is advisable. Further, providing readers with a clear understanding of how the authors are considering success, as well as working definitions of terms, will also resolve some of these challenges.

In conclusion, our paper reiterates that while AUC ROC, much like broccoli, is beneficial, it should not constitute the entirety of one's 'diet' in predictive analytics. A diversified approach, employing multiple metrics, is crucial for a robust and effective analytics strategy. Future research should focus on developing guidelines for the combined use of these metrics, thereby enhancing the efficacy and interpretability of predictive analytics.

## 7 RECOMMENDED STATEMENTS

### 7.1 Researcher positionality statement

The authors of this paper are researchers and practitioners in the use of predictive analytics within educational technologies. As a group, the authors have written several dozen papers on specific examples of using predictive analytics, both for pure research purposes, to drive automated intervention, and to provide automated dashboards and reports to practitioners. Many of these papers have used AUC ROC, the metric discussed heavily within this current article. In addition, two of the authors have been involved in large-scale commercial projects deploying predictive analytics in education. Two authors have participated in the development of models used at scale to predict high school dropout and inform teachers and school leaders which students are at risk and key factors that could reduce drop-out risk for individual students. One of those authors has also developed models used at scale within several digital learning platforms to infer student knowledge, for use in mastery learning and/or informing instructors what skills students need to learn. The other of those authors has also developed talent analytics (job quitting prediction) models used at scale in several work sectors.

### 7.2 Ethical considerations statement

This paper does not directly involve experiments with users, deployed systems, or sensitive data. However, this paper presents a theoretical perspective regarding appropriate practice for research involving all three. The use of AUC ROC involves ethical considerations, and this article is about appropriate and ethical practice for predictive analytics work (particularly in education).

### 7.3 Adverse impact statement

This paper, in recommending appropriate uses and practices involving AUC ROC is intended to reduce adverse impacts, both from the use of AUC ROC and from substituting the use of AUC ROC with other practices that risk adverse impacts (such as using the accuracy metric instead of AUC ROC). However, there are inappropriate and problematic uses of AUC ROC, as our article acknowledges. It is impossible to guarantee that any position paper of this nature will not be used to justify practices different than those intended, although we have endeavored to write our paper in a way that carefully clarifies what it means.

## ACKNOWLEDGMENTS

We thank Chelsea Porter for her assistance in manuscript preparation.